\documentclass[letterpaper, 10 pt, conference]{ieeeconf}  
\IEEEoverridecommandlockouts                            
\usepackage{cite}
\usepackage{amsmath,amssymb,amsfonts}
\usepackage{algorithmic}
\usepackage{graphicx}
\usepackage{textcomp}
\usepackage{soul}
\usepackage{amsthm}

\allowdisplaybreaks
\usepackage{comment}
\usepackage{xcolor}
\def\BibTeX{{\rm B\kern-.05em{\sc i\kern-.025em b}\kern-.08em
    T\kern-.1667em\lower.7ex\hbox{E}\kern-.125emX}}
\begin{document}
\title{\LARGE \bf Physics-Informed Neural Network for Discovering Systems with Unmeasurable States with Application to Lithium-Ion Batteries}                                                            
\author{Yuichi Kajiura,~Jorge Espin, and Dong Zhang
\thanks{Y. Kajiura, J. Espin and D. Zhang are with the School
of Aerospace and Mechanical Engineering, The University of Oklahoma, Norman,
OK, 73019, USA. E-mail: {\tt\{yuichi, jorge.espin, dzhang\}@ou.edu.}}
}

\maketitle

\begin{abstract}
Combining machine learning with physics is a trending approach for discovering unknown dynamics, and one of the most intensively studied frameworks is the physics-informed neural network (PINN). However, PINN often fails to optimize the network due to its difficulty in concurrently minimizing multiple losses originating from the system's governing equations. This problem can be more serious when the system's states are unmeasurable, like lithium-ion batteries (LiBs). In this work, we introduce a robust method for training PINN that uses fewer loss terms and thus constructs a less complex landscape for optimization. In particular, instead of having loss terms from each differential equation, this method embeds the dynamics into a loss function that quantifies the error between observed and predicted system outputs. This is accomplished by numerically integrating the predicted states from the neural network(NN) using known dynamics and transforming them to obtain a sequence of predicted outputs. Minimizing such a loss optimizes the NN to predict states consistent with observations given the physics. Further, the system's parameters can be added to the optimization targets. To demonstrate the ability of this method to perform various modeling and control tasks, we apply it to a battery model to concurrently estimate its states and parameters. 
\end{abstract}


\section{Introduction}
\subsection{Physics-informed machine learning}
Followed by the huge success of deep learning in the last decade in many computer science applications such as image recognition and natural language processing\cite{gcb16}, there has been an increasing effort to apply deep learning to various science and engineering problems. Given the limited data availability of physical systems due to high experimental cost, the concept of fusing laws of physics into machine learning (ML) models to accelerate training, namely physics-informed machine learning (PIML), is gaining popularity. In PIML, physical laws can be integrated with machine learning through (i) leveraging known physics-based dynamics to produce training datasets, (ii) incorporating physical laws into the ML model architecture, or (iii) formulating loss functions that penalize the violation of physical laws. Among the PIML models, PINN is notably prominent, aligning with the third approach \cite{rpk19}. PINN has attracted attention within the scientific community due to its versatility. It has been effectively employed in various challenging problems, such as those with strong nonlinearities, and has demonstrated its efficacy in both solving dynamic systems (i.e., forward problems) and identifying systems' parameters (i.e., inverse problems) in many fields such as solid mechanics \cite{ema21} and fluid mechanics \cite{lhs20}. However, it is also reported that PINN often fails \cite{ft20}. Wang et al. attributed this to the numerical stiffness due to multi-scale interaction between different terms in the PINN's loss function, leading to unbalanced back-propagated gradients during model training \cite{ytp20}. 

To address this issue, a quasi-steady-state assumption (QSSA) was employed to remove stiff species from chemical kinematics before training in \cite{wwz21}. However, the non-trivial assumption (i.e., QSSA) restricts its broader applicability. More recently, initial conditions (ICs) and boundary conditions (BCs) have been directly integrated into the governing differential equations thus reducing the number of terms in the PINN's loss function in \cite{fsf22}. While this method shows its effectiveness for tackling stiff problems in which the loss terms from ICs and BCs make the optimization complex, it does not help when the loss terms from multiple differential equations create a complex loss landscape for optimization because the transformed system still has as many loss terms as the number of differential equations in the original system. 

In this paper, we propose a loss function that is calculated by a sequence of predicted outputs using numerical integration with the known dynamics. This reduces the number of loss terms when the number of outputs is smaller than the number of governing differential equations, which is often true  for a complex system in which their states are unmeasurable, like LiBs. 

\subsection{LiB state estimation and system identification}
The accelerating penetration of electric vehicles into society makes LiB one of the most urgent research topics. There is prominent research on improving or innovating LiB chemistry (e.g., solid-state LiB). On the other hand, as OEMs and battery manufacturers are building huge production lines of LiB with established chemistry, how to use existing LiBs wisely, i.e., battery management, gains increasing importance. 

The objective of battery management is to deliver the required energy and power from batteries safely and efficiently with minimized degradation. Central to this goal is the accurate estimation of the battery's internal states, such as state of charge (SoC) and state of health (SoH). However, these are challenging tasks because LiB is an inherently nonlinear system with multiple spatial scales and multi-time scale reactions. Moreover, the states are not directly measurable and must be inferred using available data. To tackle those challenges, a rich body of model-based and data-driven methods has been developed. Model-based approaches primarily rely on two types of battery models: equivalent circuit models (ECMs) and electrochemical models. ECM is a simple electric-circuit approximation that is preferred in real-time applications (i.e., embedded battery management systems) thanks to its computational efficiency \cite{p15}. On the contrary, electrochemical models, particularly one of the highest fidelity models called the pseudo two dimensional (P2D) model, use a system of nonlinear partial differential-algebraic equations to capture the spatio-temporal dynamics of lithium-ion concentrations and electric potentials \cite{dfn93}. To mitigate the intensive computational burden, reduced-order electrochemical models, such as the single particle model (SPM) and its variants, have been widely studied. Subsequently, a plethora of model-based state estimation algorithms, including Kalman filter (KF)\cite{KF}, extended KF \cite{dsf10}, unscented KF\cite{lfr20}, Luenberger observer \cite{trw15}, and sliding mode observer \cite{zdc20} have been developed. Given the time-dependent evolution of model parameters due to battery degradation, periodic parameter updates are crucial to the performance of the algorithms. For empirical ECMs, online parameter updates typically employ recursive least squares \cite{zad18}. Meanwhile, for electrochemical models, because of its high nonlinearity, offline methods based on Jacobian (e.g. Levenberg-Marquardt algorithm \cite{pgk18} and Gauss-Newton method \cite{rcb11}) or metaheuristic algorithms (e.g. genetic algorithm\cite{fms12}, particle swarm optimization\cite{rai16}, and harmony search\cite{kck19}) have been explored.
On the other hand, data-driven approaches, particularly the ones using NNs, have drawn attention due to their flexibility in capturing the inherent nonlinear relationships without requiring a deep understanding of the complex battery principles. Notable examples include a recurrent neural network(RNN) for SoC and SoH estimation \cite{ci17} and long short-term memory (LSTM) networks for remaining capacity estimation \cite{lsd21}. However, a key limitation is their reliance on cycling data with known SoC and SoH for training, which are often elusive. Additionally, pure data-driven approaches necessitate significant amounts of training data, and the resulting ``black-box" models  provide no insights into the underlying physicochemical processes or degradation mechanisms.

To overcome the individual drawbacks of model-based and data-driven approaches, there has been growing interest among the battery community in blending physics and machine learning (i.e. PIML) \cite{ayk21}. \cite{lzr21} constructed a two-dimensional grid LSTM network that maps measurable signals and unmeasurable electrochemical states generated by a P2D model. While this approach eliminates the necessity of real data for training NNs, one needs to identify all the P2D model parameters beforehand and simulate it with various operational conditions to obtain enough training data. In this paper, we show how our methodology constructs such mapping directly from measurable signals while concurrently identifying battery model parameters.

The contributions of this paper are as follows:

\begin{itemize}
    \item We propose a PINN that uses an integrator-constrained loss function. It minimizes the losses between measured signals and predicted values calculated by numerically integrating the network's outputs based on prior knowledge about the underlying physical system. Minimizing such a loss ensures the network's output aligns with the physics over the integration time frame.  For many physical systems, particularly the ones with unmeasurable states, this method ends up with fewer loss terms than the standard PINN and thus can avoid the typical failure mode of PINN due to the difficulty in optimizing multiple losses.
    \item We demonstrate the effectiveness of the proposed method using LiB as an example of a system with unmeasurable states and by performing state estimation and parameter identification. The three governing equations of the ECM model (which consists of three losses in the standard PINN implementation) are embedded into a single loss, leading to a successful optimization of the loss. The network identified the ECM parameters in training and estimated unmeasurable states in validation with satisfactory accuracy. For battery communities, to the best of the authors' knowledge, this is the first implementation of PINN that is directly trainable only with measurable signals and does not require prior identification of model parameters or preparation of training data with known states, including initial states. 
    
\end{itemize}

\begin{figure*}
\centerline{\includegraphics[width=0.8\textwidth]{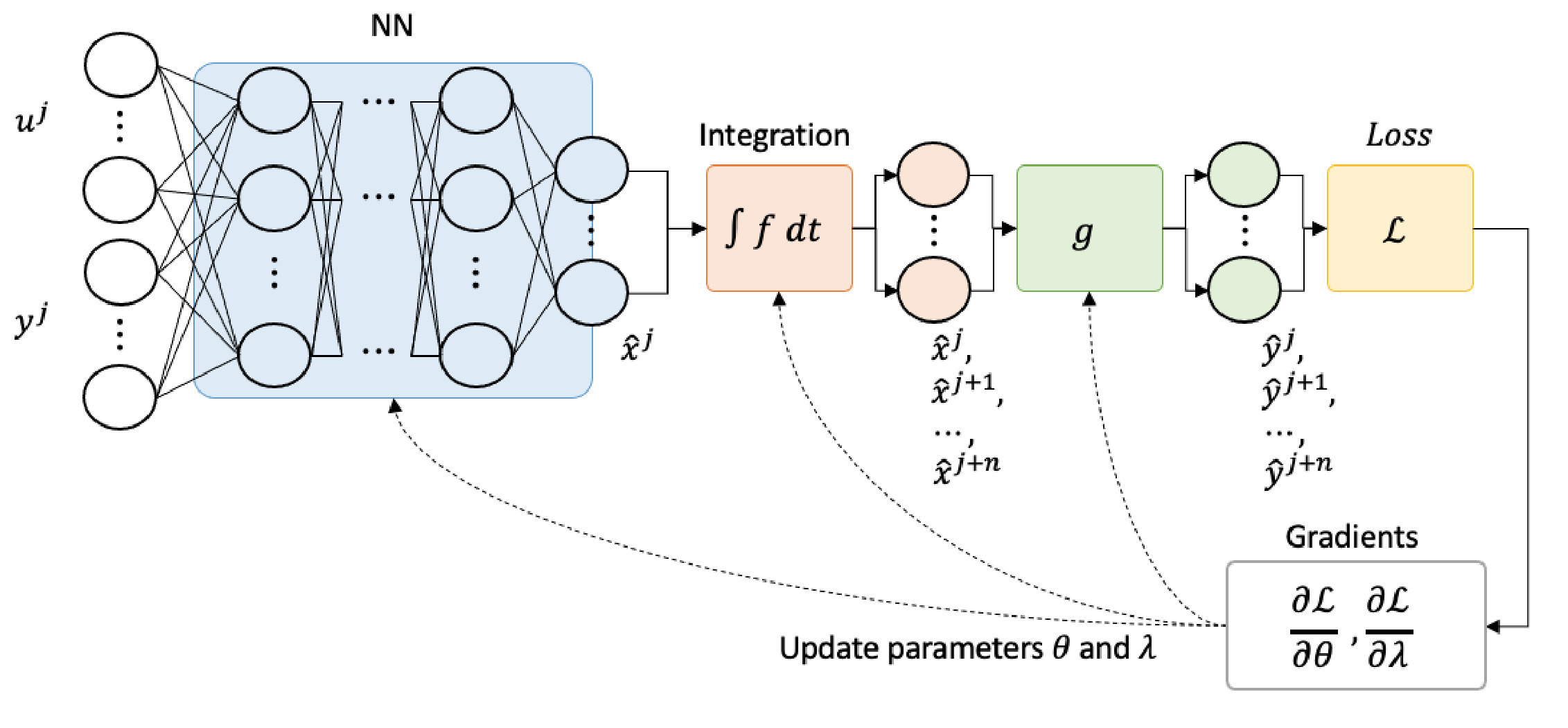}}
\caption{Framework of the PINN with integration loss}
\label{fig1}
\end{figure*}

\section{Methodology}

Consider a general form of nonlinear forced system where the states $x\in \mathbb{R}^p$ are not measurable:
\begin{align}
 \dot{x}(t) & = f(x(t), u(t); \lambda),\label{eq:f} \\
 0 & = h(x(t), u(t); \lambda), \label{eq:h} \\
 y(t) & = g(x(t), u(t); \lambda), \label{eq:g}
 \end{align}
where $t \in [0, T]$ denotes time, $x(0) = x_0$ is the initial condition which is usually not available for the system with unmeasurable states, $u(t)$ symbolizes inputs, and $y(t)\in \mathbb{R}^q$ encompasses outputs of the system. Functions $f(\cdot,\cdot; \lambda) \in \mathbb{R}^p \times \mathbb{R} \rightarrow \mathbb{R}^p$ and $g(\cdot,\cdot; \lambda)\in \mathbb{R}^p \times \mathbb{R} \rightarrow \mathbb{R}^q$ are nonlinear operators parameterized by $\lambda$, and $h(\cdot,\cdot; \lambda) \in \mathbb{R}^p \times \mathbb{R} \rightarrow \mathbb{R}^r$ can originate from ICs, BCs, or any other physical constraints. Our primary objective is to produce state estimates $\hat{x}(t)$ by a NN model using the measurable signals $u(t)$ and $y(t)$. Mathematically, 
\begin{equation}
\hat{x}(t) = \mathcal{NN}(u(t), y(t); \theta), \label{eq:nn}
\end{equation}
where $\mathcal{NN}(\cdot,\cdot; \theta)$ represents a forward pass operation of the neural network parameterized by $\theta$. One may also include historical information as inputs to the neural network to predict state estimates, as in the case of RNN\cite{ssb18}. When estimating the states at time instant $t$ using past information up to previous $\ell+1$ data points and assuming $\Delta$ as a sampling time, equation \eqref{eq:nn} can be rewritten as 
\begin{align}
\hat{x}(t) &= \mathcal{NN}(u(t),\cdots,u(t-\ell\Delta), y(t), \cdots, y(t-\ell\Delta); \theta).\label{eq:rnn}
\end{align}
Consequently, the state estimation problem is now converted to an optimization problem with respect to $\theta$. This can be solved by minimizing a loss function $\mathcal{L}$ which quantifies the prediction error. We can easily expand this to a parameter identification problem by concurrently optimizing $\lambda$ and $\theta$ during the minimization process.  The minimization task is most commonly accomplished by gradient descent\cite{GD}, while other approaches such as metaheuristic optimization are also proposed for improvements \cite{MH}.  

\subsection{Physics Informed Neural Network}

In this section, we present the loss function within the PINN framework, as outlined in \cite{rpk19}. In the subsequent section, we articulate the distinctions in our loss function for improvements. While our discussion primarily pertains to the adoption of a feedforward neural network for the NN segment of PINN, as denoted in equation \eqref{eq:nn}, the same rationale is applicable to the case of RNN, elucidated in equation \eqref{eq:rnn}. PINN originally aimed to approximate the solution to partial differential equations (PDEs). This is achieved by training a NN that takes spatial and temporal variables as inputs and predicts hidden states by minimizing a loss function. Notably, this loss function includes terms that penalize the violation of ICs, BCs, and PDE dynamics. Adapting this to our problem in \eqref{eq:f}-\eqref{eq:g} and let $\{t^j, u^j, y^j\}_{j=1}^N$ denotes training data and $\hat{x}^j$ be a simplified notion of $\hat{x}(t^j)$. Assuming there are $r$ constraints in equation \eqref{eq:h}, the PINN loss function is given by 
\begin{align}
    \label{eq:loss} 
    \mathcal{L} = \sum_{i=1}^p \omega_{f,i} \mathcal{L}_{f,i} + \sum_{i=1}^q \omega_{g,i} \mathcal{L}_{g,i} + \sum_{i=1}^r \omega_{h,i} \mathcal{L}_{h,i},
\end{align}
where 
\begin{align}
\label{eq:loss-f}
\mathcal{L}_{f, i} &= \frac{1}{N}\sum_{j=0}^{N} ( \dot{\hat{x}}_i^j - f_i(\hat{x}^j, u^j; \lambda) )^2,\\
\label{eq:loss-g}
\mathcal{L}_{g, i} &= \frac{1}{N}\sum_{j=0}^{N} ( y_i^j - g_i(\hat{x}^j, u^j; \lambda))^2\\
\label{eq:loss-h}
\mathcal{L}_{h, i} &= \frac{1}{N}\sum_{j=0}^{N} ( h_i(\hat{x}^j, u^j; \lambda) )^2.
\end{align}
In equations \eqref{eq:loss}-\eqref{eq:loss-h}, subscript $i$ denotes $i$-th element in each variable vector or function set in \eqref{eq:f}-\eqref{eq:g}. $\omega_{f,i}$, $\omega_{g, i}$ and $\omega_{h, i}$ are tunable hyper-parameters corresponding to the loss resulting from $i$-th element of respective function set. $\dot{\hat{x}}^j$ in equation \eqref{eq:loss-f} can be obtained by using automatic differentiation on $\hat{x}^j$ with respect to the NN inputs. The minimization problem based on \eqref{eq:loss} encodes model structure information into a learning algorithm to prevent the violation of the physical principles. Ultimately, the solution will avoid any non-realistic scenarios and position itself in the optimal place. However, observing equation \eqref{eq:loss}, the resulting loss function encompasses $(p + q + r)$ different losses, which clearly will impose convergence and numerical issues during optimization when the system size is relatively large. 

\subsection{Physics Informed Neural Network with Integration Loss}

Rather than concurrently minimizing the losses arising from each individual physics law, our proposed approach seamlessly embeds the differential equations with the NN. Specifically, we enforce that $\hat{x}$, upon integration through functions $f$ and subsequent transformation to $\hat{y}$ via functions $g$, aligns with the observed output $y$ over a certain duration of time. Fig.~\ref{fig1} illustrates the overall framework. Mathematically, given $\hat{x}^j$, $\hat{x}^{j+1}$ can be calculated via the integration of function $f$ over $[t^j,t^{j+1}]$, as follows, 
\begin{align} 
\hat{x}^{j+1} = \hat{x}^j + \int_{t^j}^{t^{j+1}} f(\hat{x}(\tau), u(\tau); \lambda)d\tau.  \label{eq:integ}
\end{align}
The integration can be carried out by any numerical integration method suitable for the problem, e.g., a Runge-Kutta method \cite{RK}. By iteratively performing the integration $n$ times, the following sequence can be obtained: $\{\hat{x}^j, \hat{x}^{j+1}, \cdots, \hat{x}^{j+n}\}$. Process this sequence through functions $g$, we get $\{\hat{y}^j, \hat{y}^{j+1}, \cdots, \hat{y}^{j+n}\}$.

Finally, the proposed integration-based loss function can be formulated as 
\begin{align}
\mathcal{L} &= \sum_{i=1}^q \omega_{g,i} \mathcal{L'}_{g,i} + \sum_{i=1}^r \omega_{h,i} \mathcal{L'}_{h,i}, \label{eq:loss_integ}
\end{align} 
where
\begin{align}    
\mathcal{L'}_{g, i} &= \frac{1}{n(N-n)}\sum_{j=0}^{N-n} \sum_{k=0}^{n} (y_i^{j+k} - \hat{y}_i^{j+k} )^2, \label{eq:loss_integ_g}\\
\mathcal{L'}_{h, i} &= \frac{1}{n(N-n)}\sum_{j=0}^{N-n} \sum_{k=0}^{n} ( h_i(\hat{x}^{j+k}, u^{j+k}) ;\lambda))^2.
\end{align}
Notably, equation~\eqref{eq:loss_integ} has only $(q+r)$ different losses, a reduction by $p$ compared to the loss of PINN in equation~\eqref{eq:loss}. This addresses the common failure mode of PINN stemming from unbalanced gradients from different loss terms, as elucidated in \cite{ytp20}. Furthermore, there is also a reduction of $p$ in the number of hyper-parameters to be tuned compared to equation \eqref{eq:loss}, simplifying the implementation complexity. These improvements can be significant, particularly for the system with unmeasurable states in which $p > q$. Additional constraints, such as the ones that limit the search region of model parameters, can also be integrated into the loss function for optimization.

\section{Case Study on Li-ion Battery Model}

To demonstrate the ability of the method described in the previous section, we perform a case study on state estimation and parameter identification for a lithium-ion battery model. 

\subsection{Battery Model}

\begin{figure}[t]
\centerline{\includegraphics[width=0.4\textwidth]{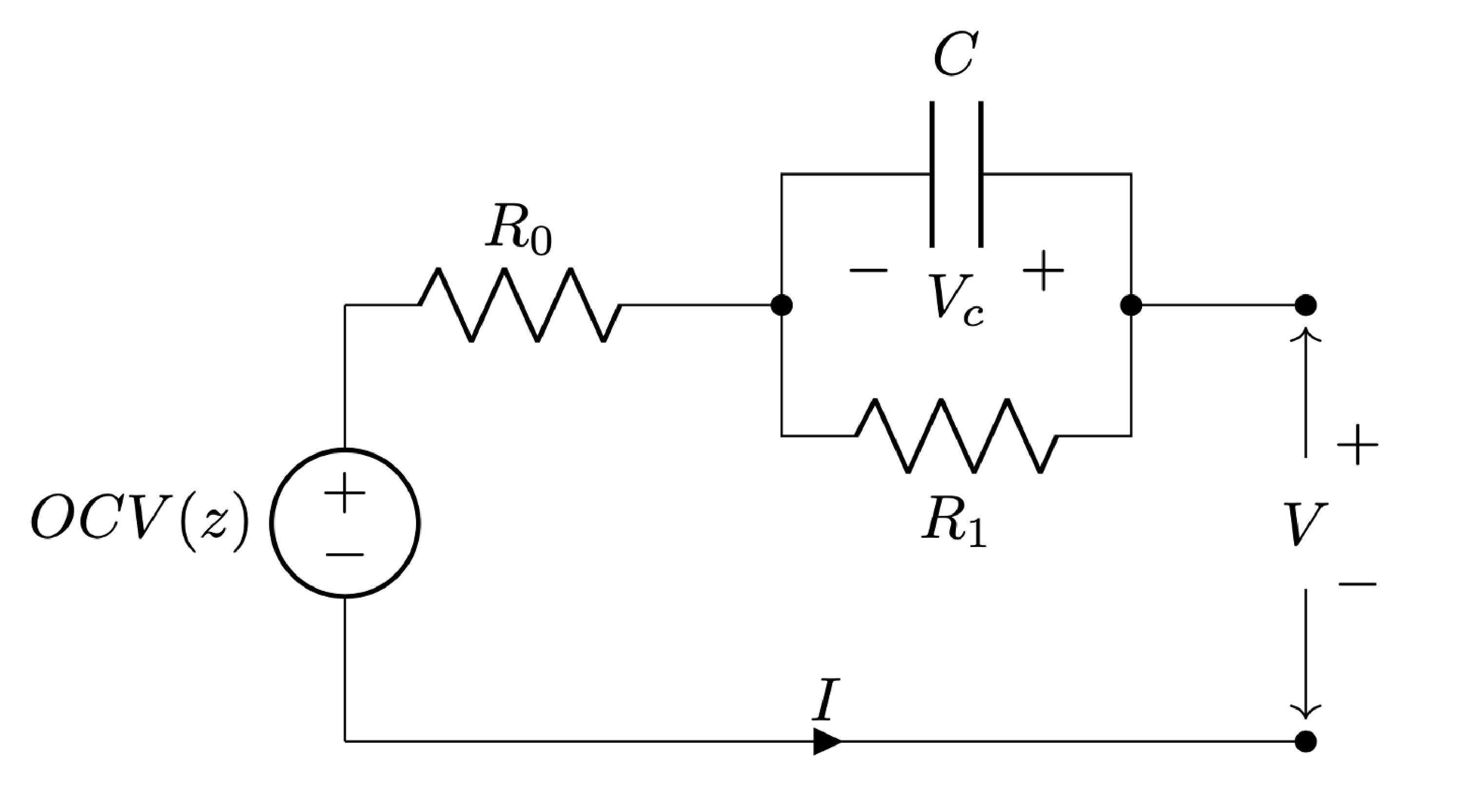}}
\caption{1 RC Pair ECM model.}
\label{fig2}
\end{figure}

For modeling a battery, we use a first-order ECM, shown in Fig.~\ref{fig2}. The model idealizes a battery using an open circuit voltage (OCV), an ohmic resistance $R_0$, and a parallel R-C pair. The system has two dynamic states, namely the state of charge $z(t)$ to represent the ratio of remaining capacity to maximum capacity and the diffusion voltage $V_c(t)$ to describe the transient behavior of the battery. The model is governed by the following set of dynamic equations: 
\begin{align}
\label{eq9}
\dot{z}(t) & = \frac{1}{3600Q}I(t), \\ 
\label{eq10}
\dot{V}_c(t) & = -\frac{1}{R_1C}V_c(t) + \frac{1}{C}I(t), \\ 
\label{eq11}
V(t) & = OCV(z(t)) + V_c(t) + R_0I(t), 
\end{align}
where $V(t)$ is the measurable terminal voltage, $I(t)$ is the input current. We use the convention to represent charging with a positive current. Furthermore, $Q$ is the total capacity in Ah, and $R_1$ and $C_1$ are resistance and capacitance to model the cell's transient behaviors. $OCV(\cdot)$ is a nonlinear function that represents the cell's equilibrium voltage as a function of $z(t)$. 
Adopting the same notation introduced in Section 2,
\begin{equation}
    x(t) = \begin{bmatrix} x_1(t) \\ x_2(t) \end{bmatrix} = \begin{bmatrix} z(t) \\ V_c(t) \end{bmatrix}, ~~u(t) = I(t), ~~y(t) = V(t),
\end{equation}
and 
\begin{equation}
f(x(t), u(t); \lambda) = \begin{bmatrix}
0 & 0\\0 & \lambda_1
\end{bmatrix} x(t) + 
\begin{bmatrix}
    \frac{1}{3600Q}\\ \lambda_2
\end{bmatrix} u(t) \label{eq:f_ecm}
\end{equation}
\begin{equation}
g(x(t), u(t); \lambda) = OCV(x_1(t)) + x_2(t) + \lambda_3 u(t) \label{eq:g_ecm}
\end{equation}
where 
\begin{equation}
    \lambda_1 = -\frac{1}{R_1C},  ~~\lambda_2 = \frac{1}{C}, ~~\lambda_3 = R_0.
\end{equation}
Our goal is to estimate unmeasurable states $z(t)$ and $V_c(t)$ while concurrently identifying system's parameters $\lambda_1$, $\lambda_2$, and $\lambda_3$ (or equivalently$R_0$, $R_1$, and $C$) only from measureable signals, $I(t)$ and $V(t)$. This is a typical SoC estimation problem under known battery capacity $Q$, but we can easily expand it to the SoH estimation problem in which $Q$ also becomes a parameter to be identified.

\subsection{NN Model}
Considering the multi-time scale nature of the LiB system, we utilize RNN as our NN model. To capture the nonlinearity between measurable signals and states, we also add a fully connected (FC) layer before the final outputs. The resulting NN is a 2 inputs - 2 outputs system that has a recurrent layer of 20 neurons followed by a FC layer of 200 neurons. The time steps that the recurrent layer considers, i.e., $\ell$ in \eqref{eq:rnn}, is set to 30. The recurrent layer used a hyperbolic tangent (Tahn) and the FC layer used a rectified linear unit (ReLU) as their activation function respectively. For numerical integration in \eqref{eq:integ}, the 4th order Runge-Kutta method was chosen, and the time steps for integration, i.e., $n$ in \eqref{eq:loss_integ_g}, is set to 30. 

\subsection{Data}
\begin{figure}[h!]
\centerline{\includegraphics[width=\linewidth]{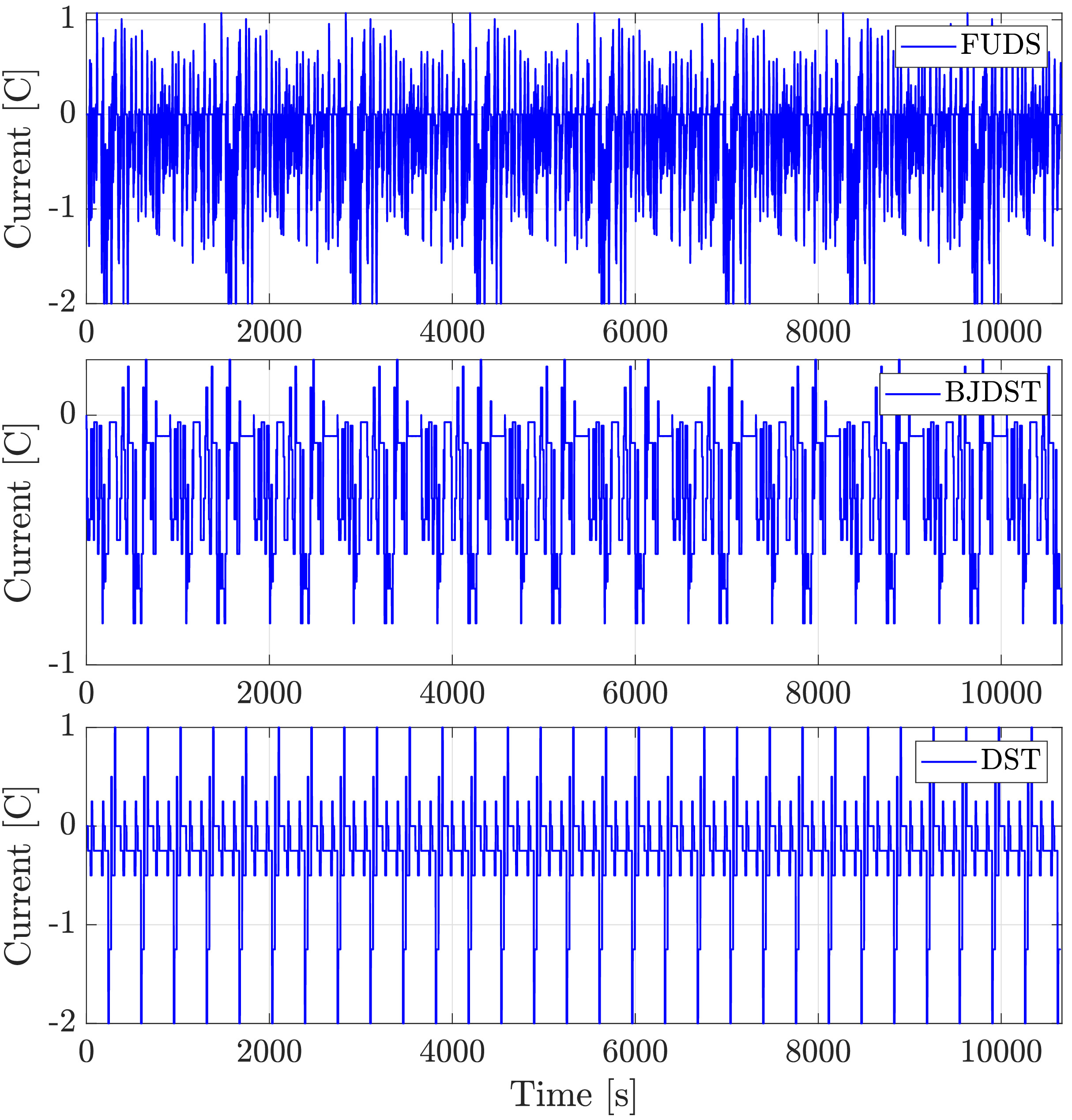}}
\caption{Current profiles used for training (FUDS and BJDST) and validation (DST).}
\label{fig3}
\end{figure}

For evaluation purposes, we use simulated data from an ECM model so that we have true states unless unmeasurable. The cell considered for the simulation is Samsung INR-18650 20R (Nominal Capacity 2.0Ah). The OCV test data using 1/20 C-rate provided by Center for Advanced Lifecycle Engineering (CALCE) \cite{b3} is used for determining $OCV(\cdot)$, where C-rate is the rate at which a battery is charged or discharged relative to its maximum capacity. The OCV curve is fitted to a 7th-order polynomial function:
\begin{equation}
    OCV(z(t)) = \sum_{m=0}^{7} a_m z^m(t).
\end{equation}
Values of $a_m$'s can be found in Table \ref{tab2} in Appendix A.
We assume that the true parameter values are $R_{0, true} = 0.06 \Omega, R_{1, true} = 0.03 \Omega , C_{true} = 1000\text{F}$, estrapolated from the value estimated in \cite{b4} and \cite{b5}. Given these parameters, the ECM model \eqref{eq9}-\eqref{eq11} is simulated with initial conditions $z(0) = 80\%$ and $V_c(0)=0$ to obtain synthetic data sets for $z(t)$, $V_c(t)$ (for validation) and $V(t)$ (for training and validation). 
Among various current profiles provided in \cite{b3}, the Federal Urban Driving Schedule (FUDS) and the Beijing Dynamic Stress Test (BJDST) are chosen for training. Dynamic cycles are selected for training the NN because they are characterized by large, non-sustained, and high-frequency currents and are useful for exploring more complete state space. In addition, those are close to the real data from on-load vehicles, and thus more suitable for evaluating the framework from a practical point of view. For validation purposes, the Dynamic Stress Test (DST) is selected. These three current profiles are shown in Fig.~\ref{fig3}

\subsection{Training}

To demonstrate the ability of the network to identify parameters during the training process, initial values for the parameter estimates are set to be $50\%$ larger than the true value ($R_{0, init} = 0.09\Omega, R_{1, init} = 0.045\Omega, C_{init} = 1500$F). Since the ECM is a single-input-single-output (SISO) system and is not subject to any algebraic constraints, the loss function \eqref{eq:loss_integ} can be further simplified to

\begin{align}
\mathcal{L} = \mathcal{L'}_g \label{eq:loss_integ_ecm},
\end{align}
To improve training efficiency, we limit the region of parameter search for $R_1$ and $C$ between 200\% and 50\% of its true values by introducing an additional constraint. We used Adam \cite{b6} for optimizing the network. The network is implemented in Python using the Pytorch framework and trained with 200k epochs. 

\subsection{Results and Discussion}

\subsubsection{Parameter Identification} After training, the error between identified $R_0$, $R_1$ and $C$ and their true values was 1\%, 4.7\%, and 5.9\% respectively, as summarized in Table 1. The smaller error of $R_0$ compared to the other two parameters aligns with the system's nature in which $R_0$ is easily identifiable. The fact that we optimized for $\lambda_1$ and $\lambda_2$ instead of $R_1$ and $C$ should also contribute to the training efficiency and accuracy, as it makes the function \eqref{eq:f_ecm} linear in the parameters. 
\begin{table}[h]
\caption{Result of parameter identification}
\begin{center}
\begin{tabular}{|c|c|c|c|}
\hline
\cline{1-4} 
\textbf{Variable [Unit]} & \textbf{True value}& \textbf{Value after training}& \textbf{Error~ [$\%$]} \\
\hline
\cline{1-4} 
$R_0$ [$\Omega$] & 0.06 & 0.0594 & 1 [$\%$] \\
\hline
$R_1$ [$\Omega$]& 0.03 & 0.0286  & 4.7 [$\%$] \\
\hline
$C$ [$F$]& 1000 & 944.1464 & 5.9 [$\%$] \\
\hline
\end{tabular}
\label{tab1}
\end{center}
\end{table}

\subsubsection{State estimation}
\begin{figure}[hb!]
\centerline{\includegraphics[width=0.97\columnwidth]{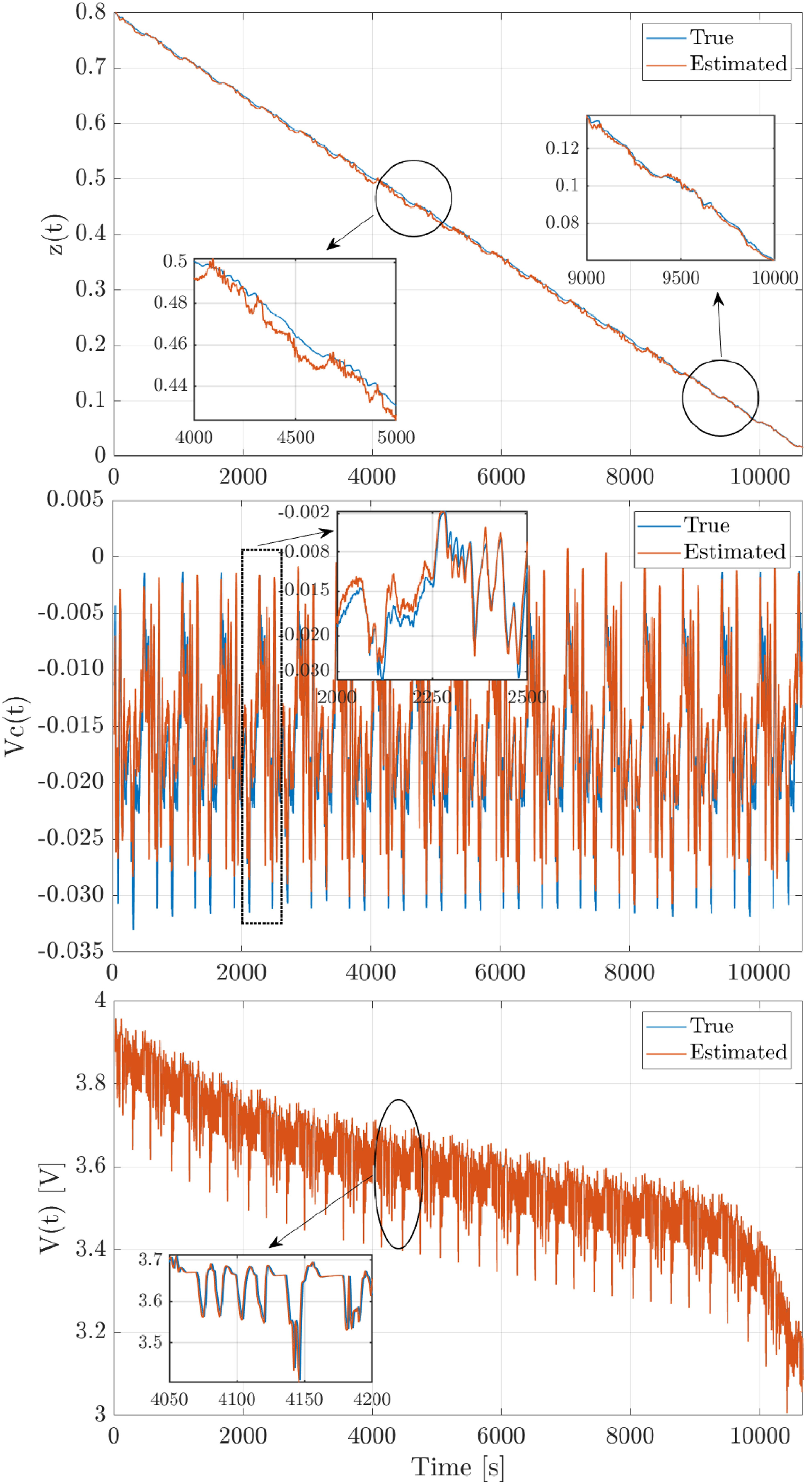}}
\caption{The results of estimated states and $V(t)$ on validation data (DST).}
\label{fig4}
\end{figure}

Using the trained NN with identified parameters, we performed the state estimation, i.e., a forward pass operation of \eqref{eq:rnn}, on validation data and then transformed $\hat{x}(t)$ by the function $g$ to obtain $\hat{V}(t)$. Fig.~\ref{fig4} shows the state estimation results. The errors of $z(t), V_c(t)$ and $V(t)$ per data point are 0.3\%, 1.9mV, and 25.1mV in average, respectively.

\subsubsection{Discussion}
As shown in the results, the proposed method achieved satisfactory accuracy for both parameter identification and state estimation problems. Notably, the training data used for those tasks only corresponds to two driving data from SoC 80\% to SoC 20\%, which can be considered as significantly smaller than the amount of data typically required for training deep learning models, realizing the original motivation of incorporating the law of physics into machine learning. The fact that we could avoid the typical failure mode of PINN can be attributed to the simplified loss function \eqref{eq:loss_integ_ecm} in which there is only one loss term and thus no competing gradients. On the other hand, the traditional PINN loss, if applied to the same ECM model, would be
\begin{align}
\mathcal{L} = \omega_{f,1}\mathcal{L}_{f,1} + \omega_{f,2}\mathcal{L}_{f,2} + \omega_g\mathcal{L}_g, 
\end{align}
where the first two terms results from \eqref{eq9} and \eqref{eq10} respectively. This would create a more complex loss landscape to optimize. The difference between the two methods could be more significant if applied to a more sophisticated electrochemical battery model that has an increased number of states.

\section{Conclusion}

A new framework for training PINN which minimizes losses calculated by numerically integrating the NN outputs is proposed. It is shown that the proposed framework results in strictly fewer terms in its loss function than the traditional PINN framework, and thus reduces the complexity in optimizing PINN. The effectiveness of this framework is then demonstrated in the state estimation and parameter identification problems for LiB. From a battery modeling and control perspective, this approach has the advantage that it does not require known parameters or states and is directly trainable only from measurable signals, posing a one-stop approach for concurrent estimation of parameters and states which has been traditionally solved by combining two methods (e.g., jacobian-based parameter search and state observer).

The limitation of this paper is that it only defines and demonstrates the framework on the system of ordinary differential equations, and on simulated data free from noise. Future work includes applying the model to a more complex system, such as electrochemical battery models, and demonstrating the effectiveness on real data with noise. For those purposes, although this paper uses a simple RNN and a popular gradient-based algorithm, i.e., Adam, for optimization, incorporation of more sophisticated NN models, such as LSTM, and optimization methods, such as metaheuristic algorithms, may need to be sought for improving training efficiency or avoiding local optima.
\appendices
\section{}
 \begin{table}[htbp]
\caption{Coefficient of estimated OCV curve}
\begin{center}
\begin{tabular}{|c|c|}
\hline
\cline{1-2} 
\textbf{Coefficient} & \textbf{Value}  \\
\hline
\cline{1-2} 
$a_0$ & 3.039475779\\
\hline
$a_1$ & 9.620312047\\
\hline
$a_2$ & - 77.31237098\\
\hline
$a_3$ & 327.4461809\\
\hline
$a_4$ & - 763.3324119\\
\hline
$a_5$ & 988.4086711\\
\hline
$a_6$ & - 662.9843922\\
\hline
$a_7$ & 179.3018624\\
\hline
\end{tabular}
\label{tab2}
\end{center}
\end{table}


\begin{thebibliography}{00}

\bibitem{gcb16} I. Goodfellow, Y. Bengio and A. Courville (2016). Deep Learning, Vol. 1. MIT press Cambridge.

\bibitem{rpk19} M. Raissi, P. Perdikaris and G.E. Karniadakis. Physics-informed neural networks: A deep learning framework for solving forward and inverse problems involving nonlinear partial differential equations, Journal of Computational Physics, vol. 378, 2019, Pages 686-707.

\bibitem{ema21}E. Haghighat, M. Raissi, A. Moure, H. Gomez and R. Juanes. A physics-informed deep learning framework for inversion and surrogate modeling in solid mechanics,
Computer Methods in Applied Mechanics and Engineering, vol. 379, 2021, 113741.

\bibitem{lhs20} L. Sun, H. Gao, S. Pan, J. Wang.
Surrogate modeling for fluid flows based on physics-constrained deep learning without simulation data,
Computer Methods in Applied Mechanics and Engineering, vol. 361, 2020, 112732.


\bibitem{ytp20} S Wang, Y Teng, P. Perdik. Understanding and mitigating gradient pathologies in physics-informed neural networks.  arXiv:2001.04536 (2020)

\bibitem{ft20} O. Fuks
H. Tchelepi. Limitations of Physics Informed Machine Learning for Nonlinear Two-Phase Transport in Porous Media. Journal of Machine Learning for Modeling and Computing, 1(1), 2020. 

\bibitem{wwz21} W. Ji, W. Qiu, Z. Shi, S Pan, S Deng. Stiff-PINN: Physics-informed neural network for stiff chemical kinetics. Journal of Physical Chemistry A, 125 (36), 2021

\bibitem{fsf22} M. De Florio, E. Schiassi, R. Furfaro. Physics-informed neural networks and functional interpolation for stiff chemical kinetics. Chaos 2022, 32, 06310

\bibitem{p15} G. Plett, \textit{Battery Management Systems: Volume I, Battery Modeling} (2015).

\bibitem{dfn93} M. Doyle, T.F. Fuller, J. Newman. Modeling of galvanostatic charge and discharge of the lithium/polymer/insertion cell. J Electrochem Soc, 140 (6) (1993), pp. 1526-1533

\bibitem{KF} C. E. Barbier,  B. Nogarede, H. L. Meyer, and S. Bensaoud. A battery state of charge indicator for electric vehicle. In Institution of Mechanical Engineers Conference Publication, vol. 5, pp. 29-29. Medical Engineering Publication LTD, 1994.

\bibitem{dsf10} D. Di Domenico, A. Stefanopoulou and G. Fiengo. Lithium-ion battery state of charge and critical surface charge estimation using an electrochemical model-based extended Kalman filter. J. Dyn. Syst. Meas. Control, vol. 132, no. 6, pp. 061302, 2010.

\bibitem{lfr20} W. Li, Y. Fan, F. Ringbeck, D. Jöst, X. Han, M. Ouyang, D. U. Sauer, Electrochemical model-based state estimation for lithium-ion batteries with adaptive unscented Kalman filter. Journal of Power Sources, vol. 476, 2020, 228534.

\bibitem{trw15}T. R. Tanim, C. D. Rahn, C. Y. Wang. State of charge estimation of a lithium ion cell based on a temperature dependent and electrolyte enhanced single particle model. Energy, vol. 80, 2015, Pages 731-739.

\bibitem{zdc20} D. Zhang, S. Dey, L. D. Couto, S. J. Moura. Battery Adaptive Observer for a Single-Particle Model With Intercalation-Induced Stress.  IEEE Transactions on Control Systems Technology, vol 28, no. 4, 2020.

\bibitem{zad18} C. Zhang, W. Allafi, Q. Dinh, P. Ascencio, J. Marco. Online estimation of battery equivalent circuit model parameters and state of charge using decoupled least squares technique. Energy, Volume 142, 2018, Pages 678-688.

\bibitem{pgk18} S. Park, D. Kato, Z. Gima, R. Klein, S. Moura. Optimal Experimental Design for Parameterization of an Electrochemical Lithium-Ion Battery Model. J Electrochem Soc, 2018, vol. 165. 

\bibitem{rcb11} V. Ramadesigan, K. Chen, N. A. Burns, V. Boovaragavan, R. D. Braatz, V. R. Subramanian. Parameter Estimation and Capacity Fade Analysis of Lithium-Ion Batteries Using Reformulated Models. J Electrochem Soc, 2011, vol. 158

\bibitem{fms12} J. C. Forman, S. J. Moura, J. L. Stein and H. K. Fathy. Genetic identification and Fisher identifiability analysis of the Doyle–Fuller–Newman model from experimental cycling of a LiFePO 4 cell. J. Power Sources, vol. 210, pp. 263-275, Jul. 2012.

\bibitem{rai16} M. A. Rahman, S. Anwar and A. Izadian. Electrochemical model parameter identification of a lithium-ion battery using particle swarm optimization method. J. Power Sources, vol. 307, pp. 86-97, Mar. 2016.

\bibitem{kck19}M. Kim, J. Chun, J. Kim, K. Kim, J. Yu, T. Kim, S. Han. Data-efficient parameter identification of electrochemical lithium-ion battery model using deep Bayesian harmony search.
Appl. Energy, 254 (2019), Article 113644

\bibitem{ci17} H. Chaoui, C.C. Ibe-Ekeocha. State of charge and state of health estimation for lithium batteries using recurrent neural networks. IEEE Trans. Veh. Technol., 66 (10) (2017), pp. 8773-8783,

\bibitem{lsd21} W. Li, N. Sengupta, P. Dechent, D. Howey, A. Annaswamy, D.U. Sauer. Online capacity estimation of lithium-ion batteries with deep long short-term memory networks. J. Power Sources, 482 (2021), p. 228863.

\bibitem{ayk21}
M. Aykol etc. Perspective—Combining Physics and Machine Learning to Predict Battery Lifetime. J Electrochem Soc, 2021, vol 168. 

\bibitem{lzr21} W. Li, J. Zhang, F. Ringbeck, D. Jöst, L. Zhang, Z. Wei, D. U. Sauer. Physics-informed neural networks for electrode-level state estimation in lithium-ion batteries. J. Power Sources, vol 506, 2021, 230034.

\bibitem{ssb18} H. Salehinejad, S. Sankar, J. Barfett, E. Colak, S. Valaee. Recent Advances in Recurrent Neural Networks. arXiv:1801.01078 (2018)

\bibitem{GD}
S Ruder. An overview of gradient descent optimization algorithms.
{\it{arXiv}}:1609.04747 (2017)

\bibitem{MH}
D. Devikanniga et al. Review of Meta-Heuristic Optimization based Artificial Neural Networks and its Applications. 2019 J. Phys.: Conf. Ser. 1362 012074

\bibitem{RK}
Epperson, James F. An introduction to numerical methods and analysis. John Wiley \& Sons, 2021.

\bibitem{b3} https://calce.umd.edu/battery-data

\bibitem{b4}
M.-K. Tran, A. DaCosta, A. Mevawalla, S. Panchal, M. Fowler. Comparative Study of Equivalent Circuit Models Performance in Four Common Lithium-Ion Batteries: LFP, NMC, LMO, NCA. Batteries 2021, 7, 51. 

\bibitem{b5}
N. Tian, Y. Wang, J. Chen, H. Fang. One-shot parameter identification of the Thevenin’s model for batteries: Methods and validation, Journal of Energy Storage, vol 29, 2020, 101282

\bibitem{b6}
D. P. Kingma, J. Ba. Adam: A Method for Stochastic Optimization. arXiv:1412.6980 (2014)


\end{thebibliography}
\end{document}